\documentclass{article}
\usepackage{ijcai21}

\usepackage{times}

\usepackage{soul}
\usepackage{url}
\usepackage[hidelinks]{hyperref}
\usepackage[utf8]{inputenc}
\usepackage[small]{caption}
\usepackage{graphicx}
\usepackage{amsmath}
\usepackage{booktabs}
\usepackage{amsmath} 
\usepackage{amssymb}  
\usepackage{amsmath,amssymb,amsfonts}
\usepackage{algorithmic}
\usepackage{graphicx}
\usepackage{textcomp}
\usepackage{xcolor}
\newcommand{\Lagr}{\mathcal{L}}
 
\usepackage{commath}
\usepackage{hyperref}
\usepackage[ruled,vlined]{algorithm2e}

\usepackage{longtable,tabularx}
\usepackage{multirow}
\usepackage{subfig}
\usepackage{wrapfig}
\def\BibTeX{{\rm B\kern-.05em{\sc i\kern-.025em b}\kern-.08em
    T\kern-.1667em\lower.7ex\hbox{E}\kern-.125emX}}

\title{Target-Following Double Deep Q-Networks for UAVs}

\author{
Sarthak Bhagat$^1$\and
P.B. Sujit$^2$
\affiliations
$^1$Department of Electronics and Communications, IIIT-Delhi, India\\
$^2$Department of Electrical Engineering and Computer Science, IISER Bhopal, India\\
\emails
{\tt sarthak16189@iiitd.ac.in},
 {\tt sujit@iiserb.ac.in}
}

\begin{document}

\maketitle

\begin{abstract}
Target tracking in unknown real-world environments in the presence of obstacles and target motion uncertainty demand agents to develop an intrinsic understanding of the environment in order to predict the suitable actions to be taken at each time step. This task requires the agents to maximize the visibility of the mobile target maneuvering randomly in a network of roads by learning a policy that takes into consideration the various aspects of a real-world environment. In this paper, we propose a DDQN-based extension to the state-of-the-art in target tracking using a UAV {\tt TF-DQN}, that we call {\tt TF-DDQN}, that isolates the value estimation and evaluation steps. Additionally, in order to carefully benchmark the performance of any given target tracking algorithm, we introduce a novel target tracking evaluation scheme that quantifies its efficacy in terms of a wide set of diverse parameters. To replicate the real-world setting, we test our approach against standard baselines for the task of target tracking in complex environments with varying drift conditions and changes in environmental configuration.
\end{abstract}

\section{Introduction}

Unmanned aerial vehicles (UAVs) have been deployed for a wide variety of complex tasks that stipulate elaborate path planning in advancing environments. Target tracking comes about as one of the tasks in which the UAV is supposed to persistently track a mobile target in the presence of multiplex structures acting as hindrances for the purpose of a diverse range of defence, surveillance and mapping applications. In general, the UAV has a limited field of view (\textit{FOV}) beyond which it does not track the target. For urban environments, this task becomes exceedingly demanding due to the presence of visibility obstruction from buildings and target motion uncertainty. Therefore, in this paper, we intend to propose a benchmark for self-adjusting path planning algorithms for target tracking in urban environments. 

Prior work on target tracking using UAVs has extensively covered a diverse range of training strategies and formulations in the last two decades. One of the most prevailing coherent approaches to target tracking is via the development of guidance laws \cite{wise2006uav,choi2014uav,oh2013rendezvous,regina2011uav,chen2009tracking,theodorakopoulos2008strategy,pothen2017curvature}. In principle, the motion model for the target should be known apriori in order to design these laws that satisfy the \textit{FOV} constraints. While these works laid the foundation for the contemporary target tracking, they failed to consider the obstacles as another additional constraint for optimization. The visibility obstruction due to obstacles in the environment urged various works \cite{shaferman2008cooperative,cook2013intelligent,yu2014cooperative} to focus on deployment of multiple cooperative UAVs in order to track a moving target persistently. In this work, we specifically prioritize our attention to the task of target tracking using only a single UAV with a limited \textit{FOV} in an environment with obstacles.

\cite{zhao2019detection} developed a YOLO-based vision algorithm to detect the target in an urban environment by supervising the motion using a simple proportional controller on an environment with the absence of any obstacles. \cite{watanabe2010optimal} proposed an optimal guidance framework that required the target motion model and was computationally expensive in practice. Additionally, \cite{ramirez2015urban} developed an information-theoretic planner that computes the estimate of the target's position using ground sensors and the UAV camera sensor. However, environments utilized in this work failed to consider the presence of obstacles and visibility constraints due to them. Despite, sufficient prior work focusing on the task of following a target using UAVs, our approach is unique in the sense that it takes account of formidable constraints of the presence of obstacles and target motion uncertainty that previous works have failed to consider simultaneously.

There have also been some prior work done in the domain of target tracking in constrained environments. \cite{theodorakopoulos2009uav} developed an iterative optimizing method to track the target in the presence of obstacles. A set of trajectories were first predicted and then evaluated based on the cost of following them, taking the obstruction and visibility constraints from obstacles into consideration. The path that minimizes the cost was determined and fed to the UAV for tracking the target. \cite{zhang2018coarse} developed a deep reinforcement technique for a camera to track the target under different aspect ratios. \cite{mueller2016benchmark} developed a UAV target tracking simulator capable of evaluating different computer vision algorithms for target detection and tracking offline accompanied by UAV tracking controllers for different target trajectories. While these works sort to perform target tracking in environments with obstacles, that constraint both the movement as well as the vision of the UAV, these environments lack urban settings that would promote development of algorithms effective in real-world scenarios. 

The task of target tracking offers a challenging engagement in which the agent is required to evaluate the goodness of each action based on the current position of the target and its own position with respect to the obstacles in the environment. We must model the agent to learn the entire structure of the environment with all its entities based on the rewards that it receives after performing a certain action and enhance its knowledge about the relative position of the target as time progresses. 
An analytical solution to this task in hand is not possible due to our prior assumption of an unknown motion model of the target vehicle. Without an analytical solution, we can not utilize a non-learning-based approach for achieving our goal. Additionally, due to the lack of availability of training data corresponding to state-action-reward triplets preceding the training stage, the only way to obtain data for learning is through interaction with the environment. Most deep learning approaches require prior training data in order to learn a network for performing the required task, therefore, former deep learning approaches directly applied to this task fail poorly.
Moreover, in our problem statement we only have the access to the simulation of the environment rather than the real one. Agent interactions with the environmental entities in real-world could be an expensive task, therefore, the agent has to be trained on simulation and later transferred to the real-world. As learning-based methods have been inadequately studied in recent literature, we sort to deploy it to perform target tracking in an evolving environmental setting.

In the past, reinforcement learning has proven to be a resourceful tool in catering to tasks that require the agent to adapt to constantly evolving environments in order to maximize its reward. In this paper, we also aim to utilize a simple reinforcement learning technique, modifying it for our specific task. For this, we extend 
a deep reinforcement learning approach called Target-Following Deep Q-Network ({\tt TF-DQN}) \cite{Bhagat2020UAVTT} to a Double DQN \cite{Hasselt2016DeepRL} that we refer to as {\tt TF-DDQN}. 
We also propose a target tracking evaluation scheme that can be utilized to quantify the performance of any given target tracking algorithm based on factors like deviation from target's trajectory, proximity to checkpoints placed along the target's trajectory, and computational resources required for training and evaluation. In general, this scheme quantifies the variant aspects of a typical target tracking system and enables us to deploy robust and efficient models in the real-world setting. This evaluation toolkit promotes further development in this domain of active research, acting as a tool to benchmark performance of diverse algorithms aiming to perform this task of target tracking.

Our contributions in this work can be summarised as follows:
\begin{itemize}
    \item We propose {\tt TF-DDQN}, a DDQN-based extension over the current state-of-the-art in target tracking {\tt TF-DQN} \cite{Bhagat2020UAVTT}, wherein we isolate the value estimation and evaluation steps in order to learn a more robust policy.
    \item We propose a standardised target tracking evaluation scheme that aids in benchmarking the performance of different algorithms over a diversified set of parameters.
    \item We provide a comprehensive evaluation of the strategies on different environmental settings including motion drift scenarios against a vision-based baseline as well as {\tt TF-DQN} \cite{Bhagat2020UAVTT}.
\end{itemize}


\section{Problem Formulation}
\label{problem_formulation}

The task of target tracking in urban environments involves a UAV capable of varying its direction of movement and altitude in order to persistently track a mobile target that is maneuvering randomly in a network of roads. We assume that the motion model of the target is probabilistic and unknown. Its movement is constrained to a grid network of roads where at each junction of road, it makes a random decision about where to direct its motion. Besides the target, the environment also consists of various obstacles of variable sizes that might cause not only collision with the UAV but may also block the vision of the target. We assume cylindrical obstacles for simplicity. Additionally, we also constraint the vision of the UAV to a circle called the field of view (\textit{FOV}) of the UAV, in which the target must lie in order for the UAV to track it.

We define the state of the UAV as its three-dimensional position in the environment given by $p_{D} = (x_{D}, y_{D}, z_{D})$. The position of target is represented in the form of two-dimensional points given by $p_{T} = (x_{T}, y_{T})$. Due to the camera resolution, we assume that the UAV can change its altitude between $z_{D} \in [h_{D}^{min}, h_{D}^{max}]$ using $n_h$ equally-spaced levels.
We also define $n \in N$ cylindrical obstacles with the $i^{th}$ obstacle having radius $r_{i}$, height $h_{i}$ with its center located at $(x_{O_{i}}, y_{O_{i}})$. We represent the entire state space and the action space of the UAV as $\mathcal{S}$ and $\mathcal{A}$ respectively.

At each time step $t$ when the agent is present at state $s_t \in \mathcal{S}$ performing an action $a_t \in \mathcal{A}$, the agent receives a reward of $r_t$ obtained from an engineered reward function based on the positions of all individual entities in the environment. The objective of the UAV is to take actions in a manner that it maximises the expected return \textit{i.e.} the expected sum of rewards which is given by $R_{t} = \sum_{\tau=1}^{\infty} \gamma^{\tau - t}r_{\tau}$. Here, $\gamma \in [0, 1] $ is the discount factor that weighs the importance of rewards in previous time steps compared to the one received most recently. In order to optimize this objective, the agent learns a policy given by $\pi(a_t|s_t,\phi)$, parameterized by $\phi$, that returns the action to be taken at each state of the agent.

\section{Proposed Approach}

In this section, we describe the key components of {\tt TF-DQN} alongside our proposed extension. We also give a elaborate description about the proposed target tracking evaluation scheme that can be used to assess the effectiveness of any given target tracking algorithm.

\subsection{Target Following Deep Q-Network ({\tt TF-DQN})}

In order to persistently track a mobile target randomly maneuvering around in a network of roads, \cite{Bhagat2020UAVTT} proposed to utilize a simple DQN to predict the value of each state-action pair for a given environment configuration. Some of the modifications proposed by this work include:
\begin{itemize}
    \item \textit{Exploration-Exploitation Tradeoff:} The authors utilized an exponentially decreasing function to compute the probability of exploration during the training procedure. This enabled the agent to explore more during the initial phases while exploit the already learnt policy during the later phases of training. This strategy is further elaborated in Appendix \ref{explore-exploit}.
    \item \textit{Piece-wise Reward Function:} A carefully engineered reward function for this specific task was utilized in order to penalise the collisions and obstructions from obstacles besides promoting persistent tracking of the target. We describe the algorithm used to obtain the rewards in Appendix \ref{reward}.
    \item \textit{Search-Space Exploration:} They proposed a novel concept of search-space wherein the actions taken by the agent were sampled from a predominantly random policy while manoeuvring in areas of uncertainty in the environment. This prevented agents from learning sub-optimal policies due to being stuck at edges of the obstacles or extremities of environment. We elaborate the motivation behind search-space exploration in Appendix \ref{search-space}.
    \item \textit{Lifelong and Curriculum Learning:} The ever-evolving nature of the environment highlights the importance of constant adaptation of the agent policy, therefore, the authors argue that the agent should never stop learning as well as test it against a curriculum learning framework for adaptation to unseen environments.
\end{itemize}

These modifications to a typical DQN enable the authors to train agents on urban environments capable of tracking moving targets in the presence of obstacles. In our extension, we also utilize these modifications and extend it to a DDQN.

\subsection{Target Following Double Deep Q-Network ({\tt {\tt TF-DDQN}})}

In a typical DQN, the value of each state-action pair is predicted using a neural network parameterized by $\theta$. This network is called the Deep Q-Network (DQN) and given by $Q(s, a; \theta)$.
To improve the stability of training, \cite{Mnih2013PlayingAW} proposed the freezing of weights of a target network with parameter $\theta^{-}$ for a fixed number of $\tau$ iteration while updating the weights of the online network by ensuring the predictions do not follow a moving target. After every $\tau$ iterations, the weights of the target network are equated with that of the online network.

The return estimation step involves the usage of a {\tt max} operator, where the expected return at a given state is the culmination of the instantaneous reward received by the agent and the expected reward at the newly obtained state.

\begin{equation}
    y_{i}^{DQN} = r + \gamma \max_{a'}Q(s', a'; \theta^{-})
    \label{q_learning_target}
\end{equation}

This operator present in the expected return estimation of a Deep Q-Learning \cite{Mnih2013PlayingAW} algorithm is often the cause of overestimation of value estimates. In order to prevent these overoptimistic values, Double Q-Learning \cite{Hasselt2015DeepRL} aims at separating the two disjoint processes of value evaluation and action selection. Similar to a DQN, a Double Deep Q-Network (DDQN) also learns two separate Q-functions, both consisting of the same skeleton network and parameterized by $\theta$ and $\theta'$ respectively. Both of these networks are updated by random allocation of experiences. 

Segregating them from Equation \ref{q_learning_target}, yields the following target:

\begin{equation}
    y_{i}^{DDQN} = r + \gamma Q(s', \underset{a} {\mathrm{argmax}}Q(s', a; \theta_t); \theta'_t)) \label{target_ddqn}
\end{equation}

Therefore, as Equation \ref{target_ddqn} suggests, we utilize the Q-function parameterized by $\theta$ for the estimation of the greedy policy. But the key difference between DDQN and DQN lies in the fact that in a DDQN, the evaluation of the value of the policy is done using another Q-function parameterized by $\theta'$. 

Therefore, applying this concept to the optimization objective of a DQN, we obtain the modified objective for a DDQN. The loss for a DDQN is given by: 

\begin{equation}
    \Lagr_{i}(\theta_{i}) = \mathbb{E}_{(s, a, r, s')\sim \mathcal{D}}\bigg[\frac{1}{2}\bigg(y_{i}^{DDQN} - Q(s, a; \theta_{i})\bigg)^{2}\bigg],
\end{equation}

and whose gradient is given by:

\begin{align}
    \nabla_{\theta_{i}} \Lagr_{i}(\theta_{i}) = \mathbb{E}_{(s, a, r, s')\sim \mathcal{D}}\bigg[\bigg(y_{i}^{DDQN} - Q(s, a; \theta_{i})\bigg) \nonumber \\ \nabla_{\theta_{i}}Q(s, a; \theta_{i})\bigg].
    \label{dqn_grad}
\end{align}

Applying all the novelties we proposed in {\tt TF-DQN} to a DDQN, we obtain a {\tt TF-DDQN} that enables our agent to persistently track down a moving target even more effectively than the {\tt TF-DQN} due to the more accurate and robust estimation and evaluation of learnt Q-function. The gradient utilized for updating the weights of these networks is obtained by replacing the target value from Equation \ref{target_ddqn} into Equation \ref{dqn_grad}.

\subsection{Target Tracking Evaluation Scheme}

We propose a set of quantitative metrics that evaluate the performance of target tracking algorithms based on a diverse range of parameters. We further divide these metrics into Generalized, Success-based, Error-based and Computation-based metrics based on their evaluation criterion.

\subsubsection{Generalized Metrics}

In order to evaluate the efficacy of our approach, we utilize a set of three generalized metrics.
\begin{itemize}
    \item \textit{\textbf{DIS}}: The average of the Euclidean distance between the target and the agent over an episode. A better trained model is expected to have a lower \textit{DIS} as the agent should be able to track the track more closely.
    \item \textit{\textbf{TIME}}: The average of time for which the target lies in the \textit{FOV} of the agent. A higher \textit{TIME} suggests the agent is able to track the target for a longer duration of time, which is desirable for our task.
    \item \textit{\textbf{REW}}: The average of the reward received by the agent over an episode. Target tracking models are supposed to have higher \textit{REW} as the agent should be able to track the target while avoiding collision and obstruction from the obstacle yielding a higher reward.
\end{itemize}

\subsubsection{Success-based Metrics}

Typical success rate used for RL applications is a percentage measure that quantifies the frequency of achieving success for a particular task. This metric in its original form does not serve the purpose for our application as for this task the agent is not only supposed to reach the goal accurately but also closely follow the target's path. Therefore, we modify success rate for our application in the following manner.

For our success-based metrics, we place checkpoints distributed along the path of the target vehicle. Our UAV is supposed to stay close to these checkpoints in order to ensure that it closely replicates the target's trajectory. Each checkpoint has a radius associated with it within which the UAV is assumed to be present at that particular checkpoint.

\begin{itemize}
    \item \textbf{\textit{Checkpoint Tracking Time (\%)}}: The percentage of time that a UAV lies within the checkpoints placed along the path. The checkpoints are densely placed for computation of this metric.
    \item \textbf{\textit{Checkpoint Tracking Success (\%)}}: The percentage of checkpoints that the UAV reaches while traversing along the trajectory. The checkpoints are sparsely placed at the corners, joints and extremities of the path of the target vehicle.
\end{itemize}

\subsubsection{Error-based Metrics}
We also propose a set of error-based metrics that compute the similarity between the target's and the UAV's trajectories based on certain distance metrics \cite{li-metrics}. As each of the four following metrics quantify the error between trajectories, we expect each of them to be as low as possible in order to closely track the target. For a set of $K$ checkpoints densely placed along the path of the target and the agent at regular intervals of time, we provide mathematical formulation of each error-based metric. The $i^{th}$ checkpoint along the agents trajectory is given by $C^{i}_{UAV}$ while that along the target's trajectory is given by $C^{i}_{T}$. We depict the $1$-norm of a vector as $||.||_{1}$.

\begin{itemize}
    \item \textbf{\textit{Root Mean Squared Error (RMSE)}}: The most commonly utilized error metric that computes the difference between two trajectories. 
    \begin{equation}
        \mathcal{E}_{RMSE} = \sqrt{ \frac{1}{K} \sum_{i=1}^{K} \Big(C^{i}_{UAV} - C^{i}_{T}\Big)^2}
    \end{equation}
    
    \item \textbf{\textit{Avg. Euclidean Error (AEE)}}:
    This metric is similar to the \textit{RMSE} metric in its formulation. Like the \textit{RMSE} metric, this metric focuses on the larger error terms and therefore, it is a pessimistic metric.
    \begin{equation}
        \mathcal{E}_{AEE} = \frac{1}{K} \sum_{i=1}^{K} \norm{\Big(C^{i}_{UAV} - C^{i}_{T}\Big)}_{1} 
    \end{equation}
    
    \item \textbf{\textit{Avg. Harmonic Error (AHE)}}:
    In contrast to the previous two error-based metrics, this one is a optimistic metric that is more dependant on the terms with a lower error.
    \begin{equation}
        \mathcal{E}_{AHE} = \frac{K}{\sum_{i=1}^{K} \norm{\Big(C^{i}_{UAV} - C^{i}_{T}\Big)}_{1}^{-1} }
    \end{equation}
    
    \item \textbf{\textit{Avg. Geometric Error (AGE)}}:
    This metric is neither optimistic nor a pessimistic metric, rather it is a balanced metric that is equally dependant on both the large as well as small error terms. It possess the characteristics of both optimistic and pessimistic metrics simultaneously.
    
    \begin{equation}
        \mathcal{E}_{AGE} = \Bigg(\prod_{i=0}^{K} \norm{\Big(C^{i}_{UAV} - C^{i}_{T}\Big)}_{1}\Bigg)^{\frac{1}{K}}
    \end{equation}
    
\end{itemize}

\subsubsection{Computation-based Metrics}

In order to quantify the computational requirements for target tracking algorithms, we propose two simple metrics that enumerate the computation training and testing time and the computation parameters utilized by the neural network. For target tracking algorithms, it is desirable to have a lower computational requirement in order for it to be deployed for real-world applications.

\begin{itemize}
    \item \textbf{\textit{Computation Time}}: Models that are required to be deployed in real-world scenarios are required to have a minimal evaluation computation time, while having a low training time is desirable for any given deep architecture. Therefore, a combination of training ($t_{tr}$) and evaluation ($t_{ev}$) time can be modeled as an evaluation metric for quantifying the feasibility of such target tracking systems.
    
    \begin{equation}
        \textbf{\textit{CT}} = t_{tr} . t_{ev}^{r}  \hspace{1cm} \text{where, } r\geq 1
    \end{equation}
    
    Here, the training time $t_{tr}$ is computed in hours while the evaluation time $t_{ev}$ is computed in seconds.
    
    \item \textbf{\textit{Computation Parameters}}: The number of parameters utilized by the entire neural network architecture can be used to quantify the computational hardware requirements of the model. A well-trained model should utilize the minimum number of parameters required for obtaining the best possible performance on the evaluation dataset. 
\end{itemize}

\section{Experiments}

In this section, we begin by describing the urban environment simulator that we utilize for all our experiments. This is followed by the details about the baselines that we use for comparison with our approach. We then demonstrate the quantitative and qualitative evaluation of our proposed approach using all the proposed metrics and trajectory visualizations. Lastly, we also evaluate the performance of our model in evolving environments using the curriculum training framework. The implementation details for {\tt TF-DQN} and {\tt TF-DDQN} are present in Appendix \ref{implementation}.

\subsection{Urban Simulator}

\begin{figure}
      \centering
      \includegraphics[scale=0.2]{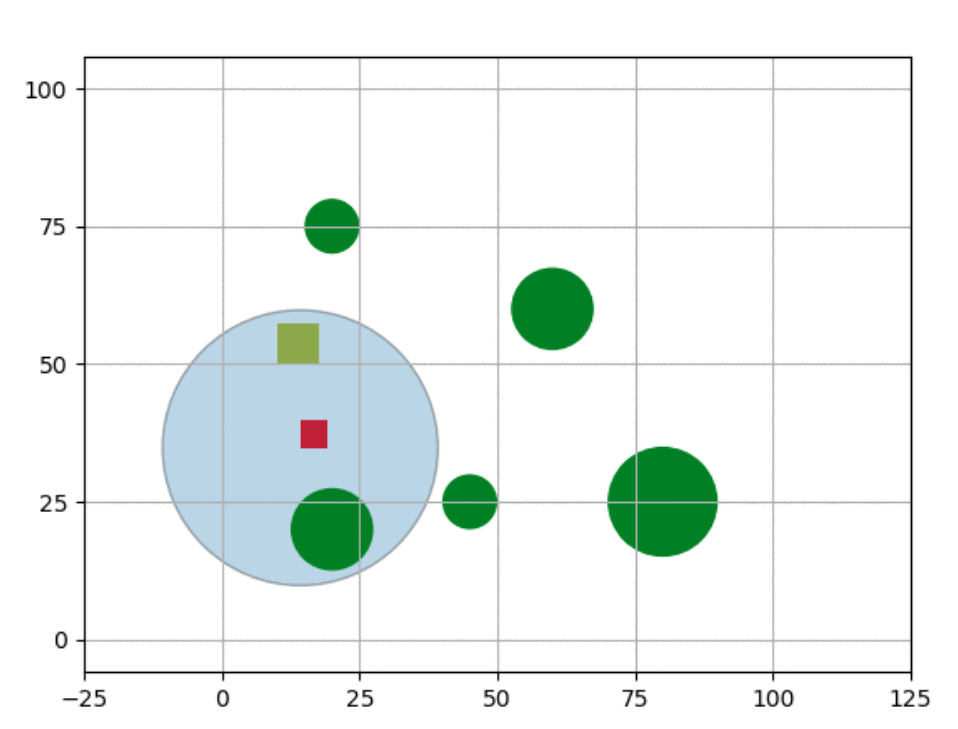}
      \caption{A screenshot of the simulator environment with red square depicting the UAV, yellow square depicting the target vehicle, green circles depicting obstacles and the large blue circle depicting the FOV of the UAV. The road network for the target vehicle to maneuver is depicted by the black lines in the figure.
      The environment available to the target is in the shape of a square with side $s = 100$.}
      \label{fig:simulation_env}
\end{figure}

In order to replicate an urban environment, we design a simulator\footnote{https://github.com/sarthak268/Target-Tracking-Simulator} that consists of a UAV with a fixed \textit{FOV}, a randomly maneuvering target vehicle and some cylindrical obstacles. A screenshot of the same is provided in Figure \ref{fig:simulation_env}.

The action space of the agent consists of a set of 6 possible actions given by $\mathcal{A} = \{{\tt north}, {\tt south}, {\tt west}, {\tt east}, {\tt up}, {\tt down}\}$, that allow it to move within its plane besides varying its altitude. The environments is of the form of a simple discretized square grid with size of each side as $s$. The agent is free to move a certain distance beyond its boundary but the target's motion is restricted to this square environment. In order to cater to camera resolution constraints, we regulate the altitude of the agent to lie within a certain fixed range, lower and higher limit of which is given by $h_D^{min}$ and $h_D^{max}$, \textit{i.e.} $h_D^{min} \leq z_D \leq h_D^{max}$. Additionally, we also discretize the altitude levels of the agent in a manner that there are $n$ equally-spaced height levels between $h_D^{min}$ and $h_D^{max}$, therefore, an increment in each height level results in an increase in height by $h_c$ given by $\frac{(h_{D}^{max} - h_{D}^{min})}{n_{h}}$.

The mobile target maneuvers in straight lines in a network of roads choosing a direction randomly at each junction of the road. We also keep the speed of the drone constant and greater than or equal to that of the target vehicle in simulation experiments. Additionally, we introduce another wind speed parameter that controls the amount of drift in the position of the UAV from the expected position when performing a particular action. The direction of this drift is arbitrary and unknown. This additional feature enables the simulator to replicate the real-world challenges and aids the assessment of our model in demanding circumstances.

\subsection{Baselines}

We compare our DDQN-based approach for target tracking against Target Following Deep Q-Network ({\tt TF-DQN}) \cite{Bhagat2020UAVTT} as well as another vision-based baseline.

\textit{Vision-based Baseline.}
We train a simple vision-based system that tries to move towards the target when it lies within its \textit{FOV} and moves randomly when it does not. This model also tries to avoid collision with obstacles visible within the \textit{FOV} of the agent by simply going past them.
Moreover, this baseline model does not try to recognize the structure of the environment and its entities, rather it attempts to persistently track the target once it has got a glance of it. Otherwise, it maneuvers randomly in the environment hoping that the target would enter its \textit{FOV} sometime. As for such a system the radius of the \textit{FOV} plays a crucial role in the performance, we trained a variety of such baseline models with varying \textit{FOV} radii for an effective comparison.
For the detection of the target within the \textit{FOV} of the UAV, we utilize Canny Edge Detection \cite{4767851} from the OpenCV\footnote{\href{https://opencv.org/}{https://opencv.org/}} library.

\subsection{Quantitative Evaluation}

We evaluate the performance of target tracking approaches against baselines using the diverse set of metrics that we propose to illustrate the efficacy of our approach with respect to a number of factors.

\subsubsection{Generalized Metric Evaluation in Variable Wind Settings.}

\begin{figure*}
\centering
\caption*{Quantitative evaluation of {\tt TF-DDQN} against other baselines in the presence of varying drift condition for different environmental settings using generalized metrics.}
\begin{minipage}{.3\textwidth}
  \centering
  \includegraphics[scale=0.41]{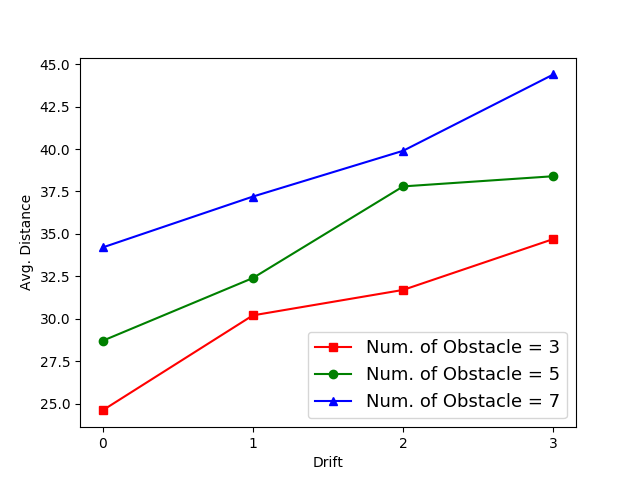}
  \caption{\textit{DIS}}
  \label{drift_distance}
\end{minipage}
\begin{minipage}{.3\textwidth}
  \centering
  \includegraphics[scale=0.41]{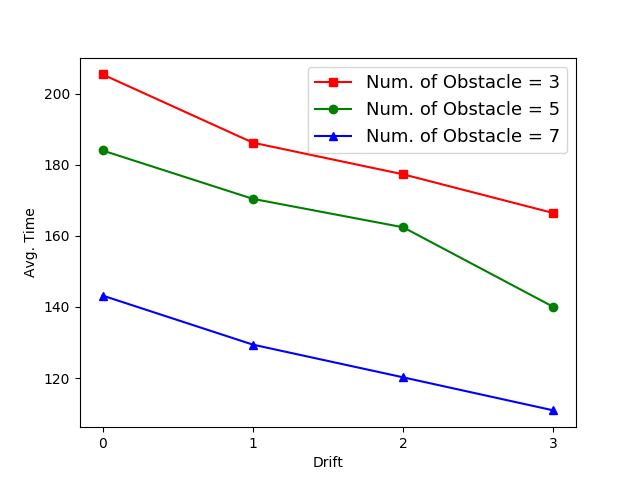}
  \caption{\textit{TIME}}
  \label{drift_time}
\end{minipage}
\begin{minipage}{.3\textwidth}
  \centering
  \includegraphics[scale=0.41]{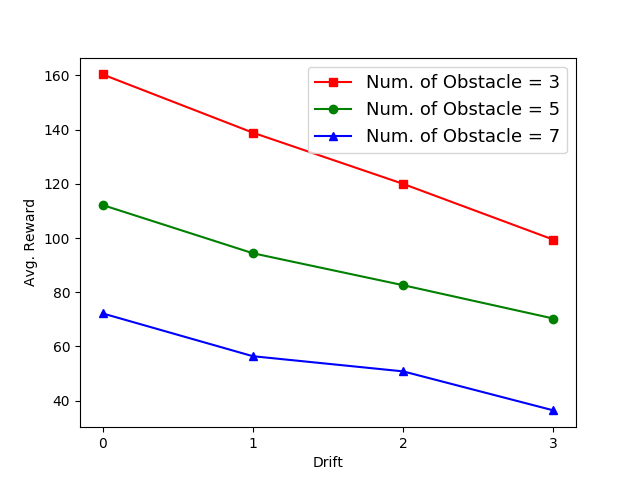}
  \caption{\textit{REW}}
  \label{drift_reward}
\end{minipage}
\end{figure*}
  
As opposed to usual setting which is completely deterministic in terms of moving to the next state as expected by the agent on performing a certain action, we also assess our model's performance in environments subjected to wind. The wind that we induce into the environment causes a drift in the motion of the agent, magnitude of which is dependant on the wind speed parameter. In order to quantify our model's performance in varying wind conditions similar to that in a real-world urban settings, we evaluate it against distinct wind speeds keeping other parameters like number, position and size of obstacles as constant. We depict the performance of our models with different wind speeds in terms of the three generalized quantitative metrics with varying environmental settings in Figure \ref{drift_distance}, \ref{drift_time} and \ref{drift_reward} respectively. 


\begin{table}
\small
    \begin{center}
        \begin{tabular}{|c|c|c|c|c|}
        \hline 
        $n$ & \textbf{Model} & \textit{DIS} $\downarrow$ & \textit{TIME} $\uparrow$ & \textit{REW} $\uparrow$ \\
        \hline 
        \multirow{3}{*}{3} & Vision Baseline & 55.2 & 80.4 & 30.8 \\
         & {\tt TF-DQN} & 31.7 & 177.3 & 120.0 \\
         & {\tt TF-DDQN} & \textbf{30.5} & \textbf{186.8} & \textbf{134.0}\\
        \hline
        \multirow{3}{*}{5} & Vision Baseline & 50.9 & 73.7 &  30.2 \\
         & {\tt TF-DQN} & 37.8 & 162.4 & 82.6 \\
        & {\tt TF-DDQN} & \textbf{32.4} & \textbf{174.5} & \textbf{90.2} \\
        \hline
        \multirow{3}{*}{7} & Vision Baseline & 56.4 & 60.8 & 22.6 \\
         & {\tt TF-DQN} & 39.9 & 120.2 & 50.8 \\
         & {\tt TF-DDQN} & \textbf{35.8} & \textbf{122.3} & \textbf{57.6} \\
       \hline
        \end{tabular}
    \end{center}
    \caption{Performance evaluation of {\tt TF-DDQN} against the baselines in the presence of static drift conditions under different environmental settings using generalized metrics.} 
    \label{table:drfit}
\end{table}

As clearly outlined in these plots, our model is able to cater to varying drift conditions, making it more suitable for the real-world setting wherein the agent is supposed to learn the relation between the action picked by the agent and the state it ends up at. Although, the performance deteriorates as the drift parameter is increased, in no situation do we observe an unanticipated instantaneous dip in the performance. 

Additionally, we also compare the performance of {\tt TF-DQN} and {\tt TF-DDQN} to the vision baseline under constant drift conditions in different environmental settings using all three generalized metrics in Table \ref{table:drfit}. Each model in depicted in this table is trained with wind parameter fixed as $2$ units.
The results depict an immensely superior performance with our approach in environments with motion drift due to the adaptable and evolving nature of the algorithm. Our model learns about the drift conditions in the environment and therefore, suggests optimal actions in a manner that it maximizes the return.
We observe a steady increase in performance of the our model by segregating the steps of evaluation and estimation via our DDQN extension. By ensuring the prevention of overestimation of value function, our model is able to learn much more efficiently and exhibits this improvement in terms of all three metrics.

\subsubsection{Success-based Metric Evaluation}

We also utilize the success-based metrics to evaluate the performance of {\tt TF-DDQN} against other baselines in terms of the success of tracking checkpoints placed along the target's path. For the computation of \textit{Tracking Time}, the placement of checkpoints is much more denser than that for measuring \textit{Tracking Success}. We place 15 checkpoints along the trajectory of the target vehicle for computing \textit{Tracking Time}, while only 10 checkpoints for measuring the \textit{Tracking Success}.

\begin{table}
\small
    \begin{center}
        \begin{tabular}{|c|c|c|}
        \hline 
        \textbf{Model} & \textit{Tracking Time (\%) $\uparrow$} & \textit{Tracking Success (\%) $\uparrow$} \\
        \hline 
        Vision Baseline & 54.6 & 40.0 \\
        {\tt TF-DQN} & 88.5 & 86.7 \\
        {\tt TF-DDQN} & \textbf{89.8} & \textbf{93.3} \\
       \hline
        \end{tabular}
    \end{center}
    \caption{Performance evaluation of {\tt TF-DDQN} against the baselines using success-based metrics.} 
    \label{table:success_metrics}
\end{table}

\subsubsection{Error-based Metric Evaluation}

We compute the error-based metrics for each of our models in order to evaluate the proximity between the trajectories of the agent and the target. Table \ref{table:error_metrics} depicts the performance of each model with respect to the error-based metrics.

\begin{table}
\small
    \begin{center}
        \begin{tabular}{|c|c|c|c|c|}
        \hline 
        \textbf{Model} & \textit{RMSE} $\downarrow$ & \textit{AEE} $\downarrow$ & \textit{AHE} $\downarrow$ & \textit{AGE} $\downarrow$ \\
        \hline 
        Vision Baseline & 48.5 & 40.4 & 36.8 & 38.6 \\
        {\tt TF-DQN} & 28.6 & 25.8 & \textbf{22.2} & 24.0 \\
        {\tt TF-DDQN} & \textbf{26.2} & \textbf{23.0} & 22.8 & \textbf{22.9} \\
       \hline
        \end{tabular}
    \end{center}
    \caption{Performance evaluation of {\tt TF-DDQN} against the baselines using error-based metrics.} 
    \label{table:error_metrics}
\end{table}

Table \ref{table:error_metrics} depict that our models is able to track the path of the target much closely than the other two baselines.

\subsubsection{Computation-based Metric Evaluation}

We also evaluate the computational requirements of our approach to validate its application in real-world scenarios. For \textit{Computation Time}, we use $r=1$.

\begin{table}
\small
    \begin{center}
        \begin{tabular}{|c|c|}
        \hline 
        \textbf{Model} & \textit{Computational Time}  \\
        \hline 
        {\tt TF-DQN} & 864 \\
        {\tt TF-DDQN} & 925 \\
       \hline
        \end{tabular}
    \end{center}
    \caption{Performance evaluation of {\tt TF-DDQN} against the baselines using computation-based metrics.} 
    \label{table:computational_metrics}
\end{table}

The DQN network utilized for both {\tt TF-DQN} and {\tt TF-DDQN} are the same, therefore, utilize the same number of parameters. Also, for the Vision Baseline, we do not utilize a neural network, therefore, it does not utilize any learnable parameters for training.
For Vision Baseline, we do not compute the \textit{Computation Time}, as there is no concept of training an agent for it. This model can directly be deployed in the environment.
As the Table \ref{table:computational_metrics} suggests, the economical evaluation time enables our model to be deployed in real world scenarios for the task of target tracking.

\subsection{Qualitative Evaluation}

We infer the performance of our model in a qualitative manner by plotting the trajectories of the our agent along with the original trajectory of the target. We test our approach on the grounds of both avoiding obstacles in the 2D plane as well as avoiding them in 3D by varying its altitude in order to follow a moving target. So as to do this, we place our agent in two disjoint environmental settings and demonstrate the effectiveness of our model in both scenarios by exhibiting the trajectories obtained.

\paragraph{Setting I} With the aim to test the ability of our model to ensure that the agent tracks the target by avoiding obstacles in the 2D plane itself, we keep the maximum height attainable by the agent lower than the height of all the obstacles.

\begin{figure*}
  \begin{center}
    \subfloat[Example 1]
    {\includegraphics[scale=0.4]{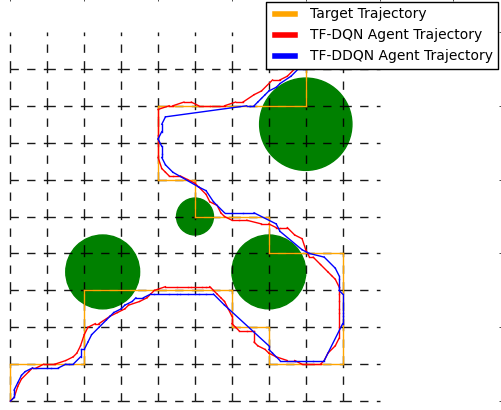}}
      \subfloat[Example 2]
      {\includegraphics[scale=0.4]{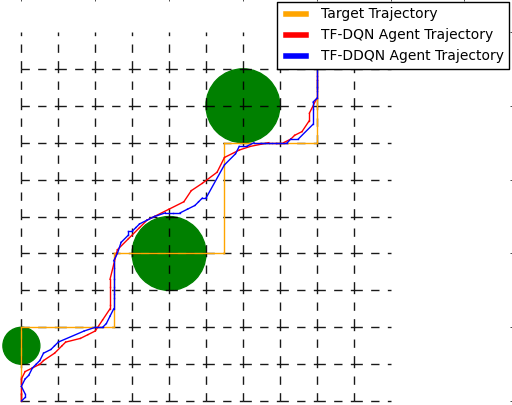}}\\
      \subfloat[Example 3]
      {\includegraphics[scale=0.4]{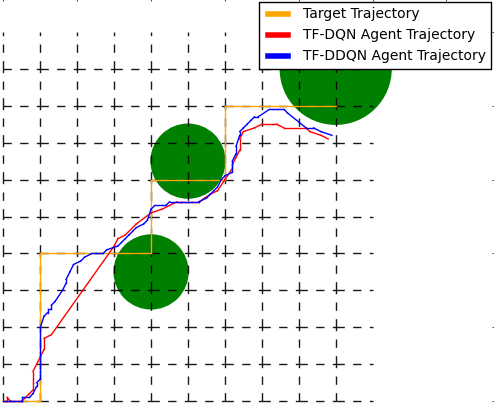}}
      \subfloat[Example 4]
      {\includegraphics[scale=0.4]{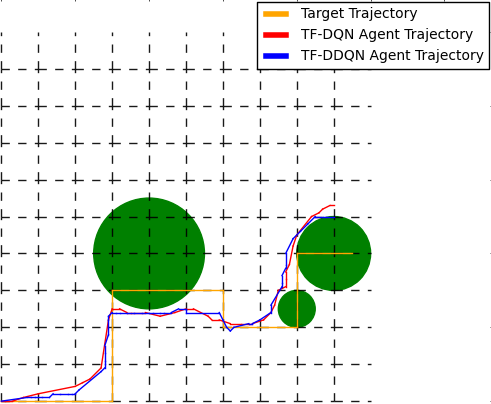}}
  \end{center}
  \caption{Trajectories of the UAV obtained by {\tt TF-DQN}, {\tt TF-DDQN} and the target vehicle when the maximum height attainable for the drone is lower than the height of any of the obstacles \textit{i.e.} $h_{D}^{max} \leq h_{i} \forall i \in n$.}
  \label{fig:trajectory_without_height}
\end{figure*}

The results depicted in Figure \ref{fig:trajectory_without_height} illustrate this scenario and demonstrate the ability of our approach to avoid obstacles with limitations on the altitude constraints to track the target persistently. The exceedingly large negative reward received on collision enforces the agent to avoid them, but this does not interfere in the prime task of following the target.

\paragraph{Setting II} As our agent is capable of varying its altitude, we would also want to evaluate the effectiveness of our agent in tracking the target by avoiding obstacles by moving over them. For this, we keep the maximum height attainable by the agent higher than the height of atleast one of the obstacles, \textit{i.e.} $h_{D}^{max} \geq h_{i}$ for some $i \in n$.

\begin{figure*}
    \centering
    \subfloat[3D Trajectory]{\label{fig:trajectory_with_height_1}  \includegraphics[scale=0.4]{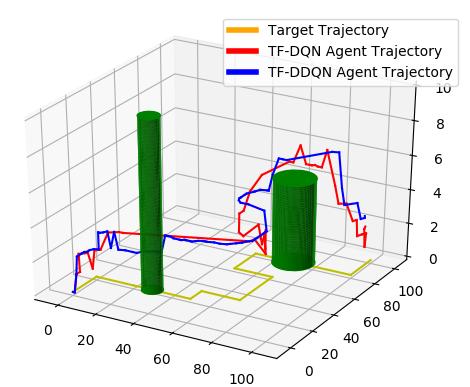}}
    \subfloat[2D Projection]{ \label{fig:trajectory_with_height_2d_1}  
    \includegraphics[scale=0.4]{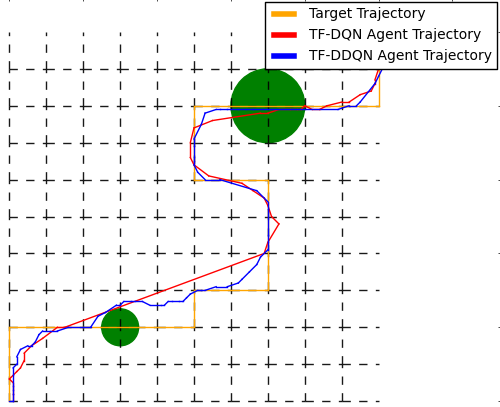}}
    \\
    \subfloat[3D Trajectory]{\label{fig:trajectory_with_height_2}  \includegraphics[scale=0.4]{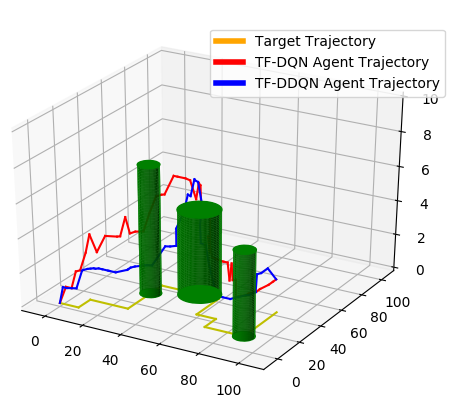}}
    \subfloat[2D Projection]{ \label{fig:trajectory_with_height_2d_2}     
    \includegraphics[scale=0.4]{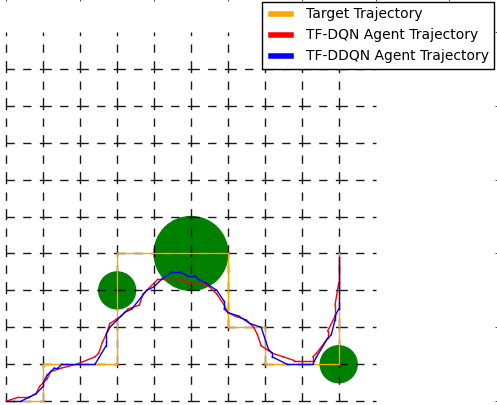}}
            \caption{(a) 3-Dimensional Trajectory of the vehicle obtained by {\tt TF-DQN} and {\tt TF-DDQN} (b) 2-dimensional projection on the x-y plane of the UAV and the target vehicle when the maximum height attainable for the drone is higher than height of atleast one of the obstacles \textit{i.e.} $h_{D}^{max} \geq h_{i}$ for some $i \in n$.}
  \end{figure*}

Using the examples \ref{fig:trajectory_with_height_1} and \ref{fig:trajectory_with_height_2} corresponding to this settings, we manifest the ability of our approach to track targets in 3D by avoiding certain obstacles by going over them while the others by going around them in 2D. The ability of our model to not only search the 2D space but also vary its altitude in order to maximize its reward, ensures that it closely replicates the real-world scenario.

As illustrated in the example provided in Figure \ref{fig:trajectory_with_height_1} and \ref{fig:trajectory_with_height_2d_1}, our agent avoids the first obstacle by going around it while the second one by going over it. Whereas in the example provided in Figure \ref{fig:trajectory_with_height_2} and \ref{fig:trajectory_with_height_2d_2}, the agent avoids the first and the last obstacle by going around it while goes over the one in the middle. In both these examples, our agent displays the ability to avoid the obstacles by utilizing the approach that yields the maximum reward and enables it to track the target as closely as possible. 
In both settings, the performance of the {\tt TF-DDQN} agent beats that of the {\tt TF-DQN} agent by a small margin by following the maneuvering target persistently using a more robust value evaluation scheme.

\subsection{Curriculum Learning Evaluation}

We also evaluate the model's performance to varying environmental settings via curriculum learning framework. For this, we first sufficiently train a network on an environment, followed by changing the environmental setting. Now, instead of completely training on the newly obtained setting, we simply finetune our trained model and evaluate its performance in terms of the same three metrics. The performance of our model on the three metrics on the original environments and on the novel environment on which the model is not thoroughly trained are provided in Table \ref{table:curriculum}. 

\begin{table}
\footnotesize
    \begin{center}
        \begin{tabular}{|c|c|c|c|}
        \hline 
        \textbf{Num. Obs, $n$} & \textit{\textbf{DIS}} $\downarrow$ & \textit{\textbf{TIME}} $\uparrow$ & \textit{\textbf{REWARD}} $\uparrow$ \\
        \hline 
        \multicolumn{4}{c}{\textbf{Vision Baseline}} \\
        \hline
        3 & 46.8 & 105.6 & 54.7 \\
        5 & 47.6 & 92.6 & 43.2 \\
        7 & 46.9 & 89.9 & 37.2 \\
        \hline
        \multicolumn{4}{c}{\textbf{{\tt TF-DQN}}}       \\
        \hline
        3 & 24.6 & 205.4 & 160.4 \\
        5 & 28.7 & 184.0 & 112.2 \\
        7 & 34.2 & \textbf{143.2} & \textbf{72.2} \\
        3 $\rightarrow$ 5 & 29.4 & 176.8 & 106.7 \\
        5 $\rightarrow$ 7 & \textbf{36.5} & 137.7 & 68.4 \\
        3 $\rightarrow$ 7 & 43.6 & \textbf{128.2} & 62.8 \\
        \hline 
        \multicolumn{4}{c}{\textbf{{\tt TF-DDQN}}}       \\
        \hline
        3 & \textbf{22.2} & \textbf{209.2} & \textbf{161.6} \\
        5 & \textbf{24.5} & \textbf{188.8} & \textbf{113.6} \\
        7 & \textbf{30.2} & \textbf{143.2} & 70.6 \\
        3 $\rightarrow$ 5 & \textbf{27.4} & \textbf{180.8} & \textbf{107.2} \\
        5 $\rightarrow$ 7 & 37.2 & \textbf{140.0} & \textbf{71.6} \\
        3 $\rightarrow$ 7 & \textbf{42.0} & 126.8 & \textbf{65.0} \\
        \hline 
        \end{tabular}
    \end{center}
    \caption{Performance of our approach on varied number on obstacles and on the task of curriculum training using generalized metrics. Here, $a$ $\rightarrow$ $b$ represents a situation wherein a model that is trained on $a$ obstacles is fine-tuned for $b$ obstacles.} \label{table:curriculum}
\end{table}

As the results in this table suggests, we obtain satisfactory performance on each of the three generalized metrics for both {\tt TF-DQN} as well as {\tt TF-DDQN} on the unseen settings even without completely training on them. The results obtained by entire training and finetuning were similar in magnitude justifying the ability of our approach to generalize to novel situations, making our model suitable for real-world scenarios.

\section{Conclusion}

In this paper, we propose a DDQN-based extension of {\tt TF-DQN} that we call {\tt TF-DDQN} wherein we separate the steps of value estimation and evaluation to avoid overestimation of values. We also propose a standardised target tracking evaluation scheme that can be used to benchmark the performance of any given target tracking algorithm based on a number of factors like proximity to target's trajectory, success rate of tracking individual checkpoints placed along target's path and computation resources required. Through a number of extensive experiments in diverse environmental settings, we conclude that our approaches outperforms current state-of-the-art in target tracking in urban environments using a UAV.


The proposed evaluation scheme provides an effective tool to assess target tracking algorithms in a range of diverse challenging environmental scenarios. This scheme promotes development in this uncharted yet rewarding domain of robotics research. The wide range of practical real-world applications of target tracking opens new doors to develop agents with enhanced computation and cognition capabilities. This work also provides the intuition behind the establishment of a wide and diverse target tracking dataset that enables models to be trained in a supervised manner for enhanced real-time performance.

\appendix

\section{Exploration-Exploitation Tradeoff} \label{explore-exploit}

Based on \cite{Bhagat2020UAVTT}, we utilize a unique approach to model the exploration-exploitation tradeoff. We utilize an exponential function that varies with number of episodes. Assuming that the agent is located at state $s_t$ at time step $t$ of $k^{th}$ episode, the action $a_t$ is chosen in the following way:

\begin{equation}
    a_{t} = \left\{
                    \begin{array}{ll}
                      rand(\mathcal{A}) & (1 - p_{sat})e^{-\alpha k} + p_{sat} \\ 
                      \underset{a} {\mathrm{argmax}} ~Q(s, a) & otherwise \\
                \end{array} 
                \right. 
\end{equation}
where $p_{sat}$ is the saturation probability and $\alpha$ is a positive constant.

\begin{figure}
  \centering
  \includegraphics[width=0.65\linewidth]{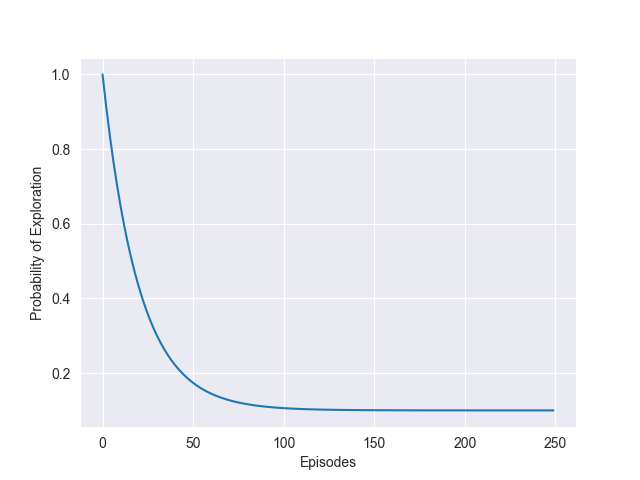}
  \captionof{figure}{Graphical plot to depict the probability of exploration over episodes.}
  \label{probability_exploration}
\end{figure}

We illustrate the probability of exploitation using a graphical plot in Figure \ref{probability_exploration} for $p_{sat} = 0.1$ and $\alpha = 2.5$ over 20 episodes. As the plot suggests, this probability decreases over episodes beginning from $1$ to saturate to a constant value of $p_{sat}$. This promotes complete exploration of the environment during the initial phases of training, aiding the performance of our model in the later stages where it has to simply pick the action corresponding to the maximum return.

\section{Designing the Piece-wise Reward Function} \label{reward}

\begin{algorithm}
\SetAlgoLined
\KwResult{Reward $r_{t}$ received by the agent at time step $t$ for picking an action $a_{t}$ at state $s_{t}$.}
\textbf{Input:} Current state of agent $p_{D}$, current state of target $p_{T}$, state of each obstacle $p_{i} \forall i \in [1, n]$, maximum allowed length of each episode $t_{max}$. Reward constants: $R_{c}$, $R_{i}$, $R_{v}^{c}$, $h_{v}^{c}$, $R_{nv}$ and $\beta$.
\\
Initialize $t$ = 0\;
 \While{$t \leq t_{max}$}{
    $collision$ = {\tt false}\;
    $intersection$ = {\tt false}\;
    \For{$i\gets1$ \KwTo $n$}{
        \eIf{$C^{i}(p_{D}, p_{i}) = 1$}{
          $collision$ = {\tt true}\;
          $break$\;
          }{}
        \eIf{$I^{i}(p_{D}, p_{T}, p_{i}) = 1$}{
          $intersection$ = {\tt true}\;
          }{}
    }
    \eIf{$collision$ = {\tt true}}{
      $t_{nv} \gets t_{nv} + 1$\;
      return $R_{c}$;
      }
      {
      \eIf{$intersection$ = {\tt true}}{
      $t_{nv} \gets t_{nv} + 1$\;
      return $R_{i}$\;
      }
      {
      \eIf{$V(p_{D}, p_{T}) = 1$}{
      $t_{nv} = 0$\;
      return $R_{v}(p_{D}, p_{T})$\;
      }
      {
      $t_{nv} \gets t_{nv} + 1$\;
      return $R_{nv}e^{-\beta t_{nv}}$\;
      }
      }
  }
  $t \gets t + 1$\;
 }
 \caption{Obtaining reward $r_{t}$ at each state-action pair trying to follow a maneuvering target.}
 \label{algo:reward}
\end{algorithm}

In order to design a custom reward function as proposed by \cite{Bhagat2020UAVTT}, we define a set of three variables that enable us to identify the objective the agent must optimize in order to perform this task effectively. We define a binary collision variable $C^{i}(p_{D}, p_{i})$ that is ${\tt true}$ (or $1$) if the UAV collides with the $i^{th}$ obstacle and ${\tt false}$ (or $0$) otherwise.

\begin{equation}
    C^{i}(p_{D}, p_{i}) =    \left\{
                    \begin{array}{ll}
                      1 & \sqrt{(x_{D} - x_{O_{i}})^{2} + (y_{D} - y_{O_{i}})^{2}} \leq r_{i}, \\ 
                      & z_{D} \leq h_{i} \\
                      0 & otherwise. \\
                \end{array} 
                \right. 
\end{equation}

In order to ensure that the visibility of the agent is not obstructed by the obstacles in the environment, we define another binary variable that corresponds to the intersection of the \textit{FOV} of the UAV directed towards the target by the obstacles. For this, we utilize the equation for the line joining the UAV to the target in 3D given by:

\begin{equation}
\label{equation_of_line}
    \frac{x - x_{D}}{x_{T} - x_{D}} = \frac{y - y_{D}}{y_{T} - y_{D}} = \frac{z - z_{D}}{- z_{D}}.
\end{equation}

The binary obstruction variable $I^{i}(p_{D}, p_{T}, p_{i})$ is ${\tt true}$ (or $1$) if the line joining the UAV and the target intersects with the $i^{th}$ obstacle and ${\tt false}$ (or $0$) otherwise. This variable $I^{i}(p_{D}, p_{T}, p_{i})$ is given by the condition for the collision of the cylinder for the $i^{th}$ obstacle with the line in 3D given by Equation \eqref{equation_of_line}.

\begin{equation}
    I^{i}(p_{D}, p_{T}, p_{i}) =    \left\{
                    \begin{array}{ll}
                      1 & \frac{z_{D}(-x_{O_{i}} + x_{D})}{x_{T} - x_{D}} + z_{D} \leq h_{i}, \\
                      & \frac{(x_{T} - x_{D})y_{O_{i}} + (y_{T} - y_{D})x_{O_{i}}}{\sqrt{(x_{T} - x_{D})^{2} + (y_{T} - y_{D})^{2}}} \leq r_{i} \\
                      0 & otherwise \\
                \end{array} 
                \right. 
\end{equation}

We also define another binary variable $V(p_{D}, p_{T})$ that represents the visibility state of the target within the \textit{FOV} of the UAV. It is ${\tt true}$ (or $1$) if the target lies in the \textit{FOV} of the UAV and ${\tt false}$ (or $0$) otherwise.

\begin{equation}
    V(p_{D}, p_{T}) = \left\{
                    \begin{array}{ll}
                      1 & \Big(x_{D} - \frac{d_{FOV}}{2}\Big) \leq  x_{T} \leq \Big(x_{D} + \frac{d_{FOV}}{2}\Big), \\
                      & \Big(y_{D} - \frac{d_{FOV}}{2}\Big) \leq  y_{T} \leq \Big(y_{D} + \frac{d_{FOV}}{2}\Big) \\
                      0 & otherwise \\
                \end{array} 
                \right.
\end{equation}

Here, the diameter of the \textit{FOV} of the UAV, $d_{FOV}$, is a function of the altitude $z_{D} \in [h_{D}^{min}, h_{D}^{max}]$ of the UAV and is given by:
\begin{equation}
    d_{FOV} = 2 z_{D} \tan(\theta_{FOV}) ,
\end{equation}

where, $\theta_{FOV}$ is the maximum possible viewing angle of the UAV calculated from the perpendicular drawn from its position to its projection on the x-y plane.

This task of target tracking requires certain constraints to be fulfilled in order for the agent to track the target closely while avoiding collision and obstruction from obstacles. Hence, our problem can be broken down into three sub-problems that include minimizing reward for collision and obstruction, maximizing reward for visibility, and penalising not being able to track the target down for a long duration of time heavily. In order to satisfy these constraints, we adopt Algorithm \ref{algo:reward} from \cite{Bhagat2020UAVTT} that takes the individual states of each individual entity as input at each time step $t$ and returns the instantaneous reward that the agent should receive. Based on this algorithm, reward $r_t$ received by the agent with state $s_t$ performing an action $a_t$ at each time step $t$ is given by:

\begin{equation}
     r_{t} = \left\{
                     \begin{array}{ll}
                       R_{c} & C^{i}(p_{D}, p_{i}) = 1; i \in [1, n]  \\ 
                       R_{i} & I^{i}(p_{D}, p_{T}, p_{i}) = 1; i \in [1, n] \\
 R_{v}(p_{D}, p_{T}) & V(p_{D}, p_{T}) = 1 \\
                       R_{nv}e^{-\beta t_{nv}} & V(p_{D}, p_{T}) = 0. \\
                 \end{array} 
                 \right.
                 \label{eq:reward_eq}
 \end{equation}
where, $R_{c}$ is the collision reward constant, $R_{i}$ is the obstruction reward constant, $R_{v}(p_{D}, p_{T})$ is the positive reward function and $R_{nv}$ is the negative reward constant. We depict the consecutive time steps for which the agent is unable to track the target as $t_{nv}$. At each time step, this variable is set to 0 if the agent views the target and is incremented when it does not.

In Equation \ref{eq:reward_eq}, we keep the value of the collision reward constant $R_c$ as a negative value with a huge magnitude to impede collision of the UAV with the obstacles. 

On training the agent, we observed several scenarios wherein it gets struck across one of the edges of the obstacles while the target is maneuvering on the other side of the obstacle, adhering to the fact that it cannot cross the obstacle due to the largely negative collision reward. In order to avoid these situations in practice, we avoid the obstruction of the vision of the agent due to the presence of obstacles by ensuring a negative reward $R_i$ for intersection of the line joining the target to the agent with any of the obstacles. 

While the target lies within the \textit{FOV} of the agent, it receives a reward given by a positive function, the value of which depends on the positions of the target and the agent. The positive reward function $R_{v}(p_{D}, p_{T})$ is given by the following function:
\begin{equation}
    R_{v}(p_{D}, p_{T}) = \frac{R_{v}^{c}}{\sqrt{(x_{D} - x_{T})^{2} + (y_{D} - y_{T})^{2}}} + \frac{h_{v}^{c}}{z_{D}},
\end{equation}
where $R_{v}^{c}$ and $h_{v}^{c}$ are both positive constants. 

In scenarios where the agent does not track the target, it receives a negative reward given by an exponential function of time steps for which it is unable to view the target consecutively. This reward is also scaled by the negative reward constant, $R_{nv}$.

\section{Search-Space Exploration} \label{search-space}

\begin{figure*}
      \centering
      \includegraphics[scale=0.75]{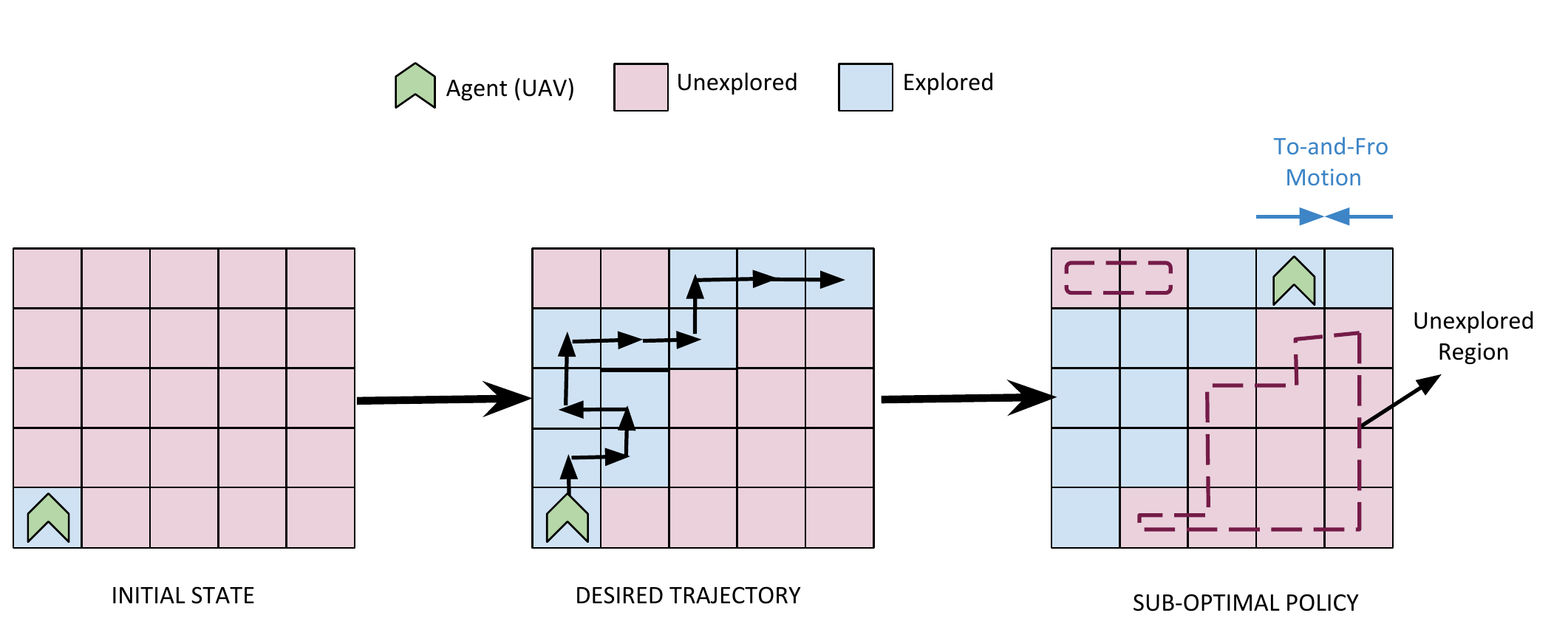}
      \caption{An illustration to depict the need of search-space exploration in target tracking agents.}
      \label{fig:search_space}
   \end{figure*}

We make use of a simple illustration to explain the concept of search-space. In Figure \ref{fig:search_space}, the left-most plot depicts an agent trying to follow a trajectory (followed by the target) in a state space represented by a simple $5\times5$ grid world. In this grid, each block represents the possible states of the agent, wherein red ones represent the unexplored ones while the blue ones represents the ones explored by the agent. For a particular episode, let us say that the agent is supposed to follow a trajectory represented by the plot in center. As it maneuvers along this trajectory, it reaches one of the extremities (top-right) of the state space. On reaching this corner, it continuously receives the same actions in the adjacent blocks at the boundary, leading to a to-and-fro motion between these two states of the agent. This recurrent motion leads to the agent getting stuck on one of the edges of the grid, leading to learning of a sub-optimal policy. In our urban simulator environment, such scenarios often arise when the agent is navigating either at the extremities of the environment or at the edges of an obstacle. In order to avoid learning of a sub-optimal policy due to lack of exploration of agent's states, \cite{Bhagat2020UAVTT} proposed search-space exploration.

Therefore, the authors modified the policy function of the agent while it lies in the search-space in a way that predominant exploration is promoted for ensuring the observation of all state-action pairs for learning the optimal action at each state.
At any moment during an episode, let us say the agent is unable to track the target for a duration of $t_{nv}$ time steps. Now, if $t_{nv}$ goes beyond a certain threshold duration of $t_{nv}^{threshold}$, \textit{\textit{i.e.}} $t_{nv} > t_{nv}^{threshold}$, we assume that the agent lies in the search-space. Therefore, for our agent with state $s_t$ at time step $t$ lying in the search-space, we define the policy function for selecting the action $a_t$ in the following way:
\begin{equation}
    a_{t} = \left\{
        \begin{array}{ll}
          rand(\mathcal{A}) & p_{SS} \\
          \underset{a} {\mathrm{argmax}} ~Q(s, a) & 1 - p_{SS} \\
    \end{array} 
    \right.
\end{equation}
where, $p_{SS}$ represents the search-space probability and is close to 1, \textit{i.e.} $p_{SS} \approx 1$ . 

\section{Implementation Details} \label{implementation}

For constructing our Q-network, we utilize a neural network with $3$ convolutional layers followed by a couple of linear fully connected layers with batch normalization after each convolutional layer. For introducing non-linearity, we use rectified linear unit (ReLU) that returns the input value for positive inputs while suppressing the negative ones. This neural networks takes the state of the agent as input and returns the action that is supposed to be taken at each time step. We list the value of each hyperparameter used while training our model in Table \ref{table:hyperparameters}.

\begin{table}
\footnotesize
    \caption{Hyperparameters used in the simulation experiments}
    \label{table:hyperparameters}
    \begin{center}
        \begin{tabular}{|c|c|c|}
        \hline
        \textbf{Hyperparameter Name} & \textbf{Symbol} & \textbf{Range} \\
        \hline
        Collision Reward Constant & $\|R_{c}\|$ & 1000-2000 \\
        \hline
        Intersection Reward Constant & $\|R_{i}\|$ & 30-100 \\
        \hline
        Positive Reward Distance Constant & $R_{v}^{c}$ & 3000-4500 \\
        \hline 
        Positive Reward Height Constant & $h_{v}^{c}$ & 1500-5000 \\
        \hline
        Negative Reward Constant & $\|R_{nv}\|$ & 1-50 \\
        \hline
        Time Constant in Negative Reward & $\beta$ & 1-10 \\
        \hline 
        Episode Constant in Action Selection & $\alpha$ & 0.1-5 \\
        \hline
        Saturation Probability & $p_{sat}$ & 0.1-0.4 \\
        \hline
        Search-Space Probability & $p_{SS}$ & 0.9-0.95 \\
        \hline 
        Min. Attainable Height for UAV & $h_{min}$ & 1-10 \\
        \hline
        Max. Attainable Height for UAV & $h_{max}$ & 10-60 \\
        \hline
        Threshold Steps for entering Search-Space & $t_{nv}^{threshold}$ & 3-10 \\
        \hline
        No. of Obstacles in the Environment & $n$ & 2-7 \\
        \hline
        Side of the Square of Environment & $s$ & 100-200 \\
        \hline
        No. of Possible Height Levels for UAV & $n_{h}$ & 5-20 \\
        \hline
        Height of obstacles & $h_{i} \forall i \in n$ & 1-50 \\
        \hline
        Height Constant & $h_{c}$ & 1-10 \\
        \hline
        Radius of obstacles & $r_{i} \forall i \in n$ & 2.5-10 \\ \hline
        Max. Viewing Angle of UAV & $\theta_{FOV}$ & $30^\text{o}-45^\text{o}$ \\ \hline
        Max. length of an episode & $t_{max}$ & 500 \\
        \hline
        Discount Factor in Return Calculation & $\gamma$ & 0.1 \\
        \hline
        Learning Rate in Gradient Descent & $\alpha_{LR}$ & 0.01 \\
        \hline
        \end{tabular}
    \end{center}
\end{table}

\bibliographystyle{named}
\bibliography{main}

\begin{thebibliography}{}

\bibitem[\protect\citeauthoryear{{Bhagat} and {Sujit}}{2020}]{Bhagat2020UAVTT}
S.~{Bhagat} and P.~B. {Sujit}.
\newblock {UAV} {T}arget {T}racking in {U}rban {E}nvironments using {D}eep
  {R}einforcement {L}earning.
\newblock In {\em 2020 International Conference on Unmanned Aircraft Systems
  (ICUAS)}, pages 694--701, 2020.

\bibitem[\protect\citeauthoryear{{Canny}}{1986}]{4767851}
J.~{Canny}.
\newblock A computational approach to edge detection.
\newblock {\em IEEE Transactions on Pattern Analysis and Machine Intelligence},
  PAMI-8(6):679--698, 1986.

\bibitem[\protect\citeauthoryear{Chen \bgroup \em et al.\egroup
  }{2009}]{chen2009tracking}
Hongda Chen, KuoChu Chang, and Craig~S Agate.
\newblock Tracking with {UAV} using {T}angent-plus-{L}yapunov {V}ector {F}ield
  {G}uidance.
\newblock In {\em 2009 12th International Conference on Information Fusion},
  pages 363--372. IEEE, 2009.

\bibitem[\protect\citeauthoryear{Choi and Kim}{2014}]{choi2014uav}
Hyunjin Choi and Youdan Kim.
\newblock {UAV} {G}uidance using a {M}onocular-{V}ision {S}ensor for {A}erial
  {T}arget {T}racking.
\newblock {\em Control Engineering Practice}, 22:10--19, 2014.

\bibitem[\protect\citeauthoryear{Cook \bgroup \em et al.\egroup
  }{2013}]{cook2013intelligent}
Kevin Cook, Everett Bryan, Huili Yu, He~Bai, Kevin Seppi, and Randal Beard.
\newblock Intelligent {C}ooperative {C}ontrol for {U}rban {T}racking with
  {U}nmanned {A}ir {V}ehicles.
\newblock In {\em 2013 International Conference on Unmanned Aircraft Systems
  (ICUAS)}, pages 1--7. IEEE, 2013.

\bibitem[\protect\citeauthoryear{Hasselt \bgroup \em et al.\egroup
  }{2016}]{Hasselt2016DeepRL}
H.~V. Hasselt, A.~Guez, and D.~Silver.
\newblock Deep reinforcement learning with double q-learning.
\newblock {\em ArXiv}, abs/1509.06461, 2016.

\bibitem[\protect\citeauthoryear{Li and Zhao}{2006}]{li-metrics}
X.~Li and Zhanlue Zhao.
\newblock Evaluation of estimation algorithms. part 1: Incomprehensive measures
  of performance.
\newblock {\em Aerospace and Electronic Systems, IEEE Transactions on}, 42:1340
  -- 1358, 11 2006.

\bibitem[\protect\citeauthoryear{Mnih \bgroup \em et al.\egroup
  }{2013}]{Mnih2013PlayingAW}
Volodymyr Mnih, Koray Kavukcuoglu, David Silver, Alex Graves, Ioannis
  Antonoglou, Daan Wierstra, and Martin~A. Riedmiller.
\newblock Playing {A}tari with {D}eep {R}einforcement {L}earning.
\newblock {\em ArXiv}, abs/1312.5602, 2013.

\bibitem[\protect\citeauthoryear{Mueller \bgroup \em et al.\egroup
  }{2016}]{mueller2016benchmark}
Matthias Mueller, Neil Smith, and Bernard Ghanem.
\newblock {A} {B}enchmark and {S}imulator for {UAV} {T}racking.
\newblock In {\em European conference on computer vision}, pages 445--461.
  Springer, 2016.

\bibitem[\protect\citeauthoryear{Oh \bgroup \em et al.\egroup
  }{2013}]{oh2013rendezvous}
Hyondong Oh, Seungkeun Kim, Hyo-Sang Shin, Brian~A White, Antonios Tsourdos,
  and Camille~Alain Rabbath.
\newblock Rendezvous and {S}tandoff {T}arget {T}racking {G}uidance using
  {D}ifferential {G}eometry.
\newblock {\em Journal of Intelligent \& Robotic Systems}, 69(1-4):389--405,
  2013.

\bibitem[\protect\citeauthoryear{Pothen and Ratnoo}{2017}]{pothen2017curvature}
Abin~Alex Pothen and Ashwini Ratnoo.
\newblock Curvature-constrained {L}yapunov {V}ector {F}ield for {S}tandoff
  {T}arget {T}racking.
\newblock {\em Journal of Guidance, Control, and Dynamics}, 40(10):2729--2736,
  2017.

\bibitem[\protect\citeauthoryear{Ramirez-Paredes \bgroup \em et al.\egroup
  }{2015}]{ramirez2015urban}
J-P Ramirez-Paredes, Emily~A Doucette, J~Willard Curtis, and Nicholas~R Gans.
\newblock {U}rban {T}arget {S}earch and {T}racking using a {UAV} and
  {U}nattended {G}round {S}ensors.
\newblock In {\em 2015 American Control Conference (ACC)}, pages 2401--2407.
  IEEE, 2015.

\bibitem[\protect\citeauthoryear{Regina and Zanzi}{2011}]{regina2011uav}
Niki Regina and Matteo Zanzi.
\newblock {UAV} {G}uidance {L}aw for {G}round-based {T}arget {T}rajectory
  {T}racking and {L}oitering.
\newblock In {\em 2011 Aerospace Conference}, pages 1--9. IEEE, 2011.

\bibitem[\protect\citeauthoryear{Shaferman and
  Shima}{2008}]{shaferman2008cooperative}
Vitaly Shaferman and Tal Shima.
\newblock Cooperative {UAV} {T}racking under {U}rban {O}cclusions and
  {A}irspace {L}imitations.
\newblock In {\em AIAA Guidance, Navigation and Control Conference and
  Exhibit}, page 7136, 2008.

\bibitem[\protect\citeauthoryear{Theodorakopoulos and
  Lacroix}{2008}]{theodorakopoulos2008strategy}
Panagiotis Theodorakopoulos and Simon Lacroix.
\newblock A {S}trategy for {T}racking a {G}round {T}arget with a {UAV}.
\newblock In {\em 2008 IEEE/RSJ International Conference on Intelligent Robots
  and Systems}, pages 1254--1259. IEEE, 2008.

\bibitem[\protect\citeauthoryear{Theodorakopoulos and
  Lacroix}{2009}]{theodorakopoulos2009uav}
Panagiotis Theodorakopoulos and Simon Lacroix.
\newblock {UAV} {T}arget {T}racking using an {A}dversarial {I}terative
  {P}rediction.
\newblock In {\em 2009 IEEE International Conference on Robotics and
  Automation}, pages 2866--2871. IEEE, 2009.

\bibitem[\protect\citeauthoryear{van Hasselt \bgroup \em et al.\egroup
  }{2015}]{Hasselt2015DeepRL}
Hado van Hasselt, Arthur Guez, and David Silver.
\newblock Deep {R}einforcement {L}earning with {D}ouble {Q}-{L}earning.
\newblock In {\em AAAI}, 2015.

\bibitem[\protect\citeauthoryear{Watanabe and
  Fabiani}{2010}]{watanabe2010optimal}
Yoko Watanabe and Patrick Fabiani.
\newblock {O}ptimal {G}uidance {D}esign for {UAV} {V}isual {T}arget {T}racking
  in an {U}rban {E}nvironment.
\newblock {\em IFAC Proceedings Volumes}, 43(15):69--74, 2010.

\bibitem[\protect\citeauthoryear{Wise and Rysdyk}{2006}]{wise2006uav}
Richard Wise and Rolf Rysdyk.
\newblock {UAV} {C}oordination for {A}utonomous {T}arget {T}racking.
\newblock In {\em AIAA Guidance, Navigation, and Control Conference and
  Exhibit}, page 6453, 2006.

\bibitem[\protect\citeauthoryear{Yu \bgroup \em et al.\egroup
  }{2014}]{yu2014cooperative}
Huili Yu, Kevin Meier, Matthew Argyle, and Randal~W Beard.
\newblock Cooperative {P}ath {P}lanning for {T}arget {T}racking in {U}rban
  {E}nvironments using {U}nmanned {A}ir and {G}round {V}ehicles.
\newblock {\em IEEE/ASME Transactions on Mechatronics}, 20(2):541--552, 2014.

\bibitem[\protect\citeauthoryear{Zhang \bgroup \em et al.\egroup
  }{2018}]{zhang2018coarse}
Wei Zhang, Ke~Song, Xuewen Rong, and Yibin Li.
\newblock {C}oarse-to-fine {UAV} {T}arget {T}racking with {D}eep
  {R}einforcement {L}earning.
\newblock {\em IEEE Transactions on Automation Science and Engineering},
  16(4):1522--1530, 2018.

\bibitem[\protect\citeauthoryear{Zhao \bgroup \em et al.\egroup
  }{2019}]{zhao2019detection}
Xiaoyue Zhao, Fangling Pu, Zhihang Wang, Hongyu Chen, and Zhaozhuo Xu.
\newblock {D}etection, {T}racking, and {G}eolocation of {M}oving {V}ehicle from
  {UAV} {U}sing {M}onocular {C}amera.
\newblock {\em IEEE Access}, 7:101160--101170, 2019.

\end{thebibliography}

\end{document}